\documentclass[runningheads]{llncs}
\usepackage[T1]{fontenc}
\usepackage{graphicx,verbatim}
\newcommand*\samethanks[1][\value{footnote}]{\footnotemark[#1]}
\usepackage{caption}
\usepackage{amsmath}
\usepackage{amssymb}
\usepackage[table]{xcolor}
\usepackage{rotating}
\usepackage{xcolor}
\usepackage[T1]{fontenc}
\usepackage[utf8]{inputenc}
\definecolor[named]{ACMDarkBlue}{cmyk}{1,0.58,0,0.21}

\usepackage[utf8]{inputenc}
\usepackage[colorlinks]{hyperref}
\usepackage{color}
\hypersetup{colorlinks, allcolors=black}

\usepackage{enumitem}

\begin{document}

\title{CytoSAE: Interpretable Cell Embeddings for Hematology}
 
\author{Muhammed Furkan Dasdelen\inst{1}%
\and
Hyesu Lim\inst{2,3}%
\and\\
Michele Buck\inst{4}%
\and
Katharina S. Götze\inst{4}%
\and\\
Carsten Marr\inst{1}%
\thanks{Co-corresponding: \email{\{carsten.marr,steffen.schneider\}@helmholtz-munich.de}.\\
}\and
Steffen Schneider\inst{2,5}%
\samethanks}

\authorrunning{Dasdelen et al.}

\institute{%
    Institute of AI for Health, Helmholtz Munich \and
    Institute of Computational Biology, Helmholtz Munich \and
    Korea Advanced Institute of Science \& Technology \and
    Medical Department for Hematology and Oncology, Technical University Munich \and
    Munich Center for Machine Learning (MCML)
}
    
\maketitle              %
\begin{abstract}
    Sparse autoencoders (SAEs) emerged as a promising tool for mechanistic interpretability of transformer-based foundation models. Very recently, SAEs were also adopted for the visual domain, enabling the discovery of visual concepts and their patch-wise attribution to tokens in the transformer model. While a growing number of foundation models emerged for medical imaging, tools for explaining their inferences are still lacking. In this work, we show the applicability of SAEs for hematology. We propose CytoSAE, a sparse autoencoder which is trained on over 40,000 peripheral blood single-cell images. CytoSAE generalizes to diverse and out-of-domain datasets, including bone marrow cytology, where it identifies morphologically relevant concepts which we validated with medical experts. Furthermore, we demonstrate scenarios in which CytoSAE can generate patient-specific and disease-specific concepts, enabling the detection of pathognomonic cells and localized cellular abnormalities at the patch level.
    We quantified the effect of concepts on a patient-level AML subtype classification task and show that CytoSAE concepts reach performance comparable to the state-of-the-art, while offering explainability on the sub-cellular level.
    Source code and model weights are available at \url{https://github.com/dynamical-inference/cytosae}.
\end{abstract}
\section{Introduction}

The advancement of foundation models has considerably popularized the application of machine learning models in various domains of medical imaging, including histopathology and cytology \cite{moor2023foundation,chen2024towards,koch2024dinobloom}. These models have demonstrated remarkable capabilities in tasks such as disease classification, cell segmentation, and feature extraction \cite{koch2024dinobloom,chen2024towards}. However, despite their high performance, they largely remain black-box systems.
This lack of transparency challenges reliable clinical adoption, where explainability is critical for ensuring trust in AI-driven diagnostics \cite{moor2023foundation} and has legal relevance \cite{EU_AI_Act_2024}.

To improve the explainability of deep learning models in medical imaging, various methods have been introduced \cite{chen2022explainable}, including Class Activation Maps (CAMs) \cite{zhou2016learning} and attention-based approaches such as attention rollout \cite{abnar2020quantifying,wagner2023transformer}. Recently, weakly supervised multiple instance learning (MIL) methods have been applied in hematology, similar to their use in histology, enabling single-cell level explainability by identifying disease-related cells \cite{hehr2023explainable}. These techniques generate heatmaps or weighted attention scores that highlight relevant regions or key instances within input bags.  While these methods offer some level of insight into model predictions, they often lack fine-grained attribution and do not provide a structured understanding of the morphological concepts that contribute to a prediction. Sparse dictionary learning has emerged as a promising approach for enhancing neural network interpretability \cite{elhage2021mathematical,huben2023sparse,bricken2023monosemanticity,sharkey2025open,templeton2024scaling,rajamanoharan2024improving,makhzani2013k,gao2024scaling}. This method decomposes network representations into a set of small, independent components that capture distinct morphological features, such as cell shape, staining intensity, or nuclear segmentation.
A sparse autoencoder (SAE) is a two-layer neural network designed for unsupervised decomposition of high-dimensional representations into sparse, interpretable components. %
SAEs have been successfully applied in the context of language~\cite{bricken2023monosemanticity,lieberum2024gemma,marks2024sparse} and computer vision~\cite{lim2024sparse,stevens2025sparse,hugo2024,cywinski2025saeuron}. Although attempts have been made to extend its application to medical imaging, such as X-Ray \cite{abdulaal2024x} or pathology \cite{lelearning}, their applicability to hematology remains to be demonstrated.

In this work, we introduce CytoSAE, a sparse autoencoder designed for morphological concept discovery in hematological imaging. CytoSAE is trained on embeddings from DinoBloom-B \cite{koch2024dinobloom}, a foundation model for hematology, and learns latent morphological concepts that generalize across diverse datasets, including peripheral blood smears and bone marrow cytology. We validate the discovered concepts with expert annotations.
To demonstrate the utility of CytoSAE, we applied it to acute myeloid leukemia (AML) subtyping data, generating patient-level concept activations (``barcodes'') and aggregating these concepts at the disease level. Our analysis shows that disease-specific concept distributions can identify morphological hallmarks of AML subtypes, providing a new level of interpretability for AI-driven hematological diagnostics.

\section{CytoSAE: Concept discovery for hematology imaging}
\label{sec:2}
\begin{figure}[t]
\includegraphics[width=\textwidth]{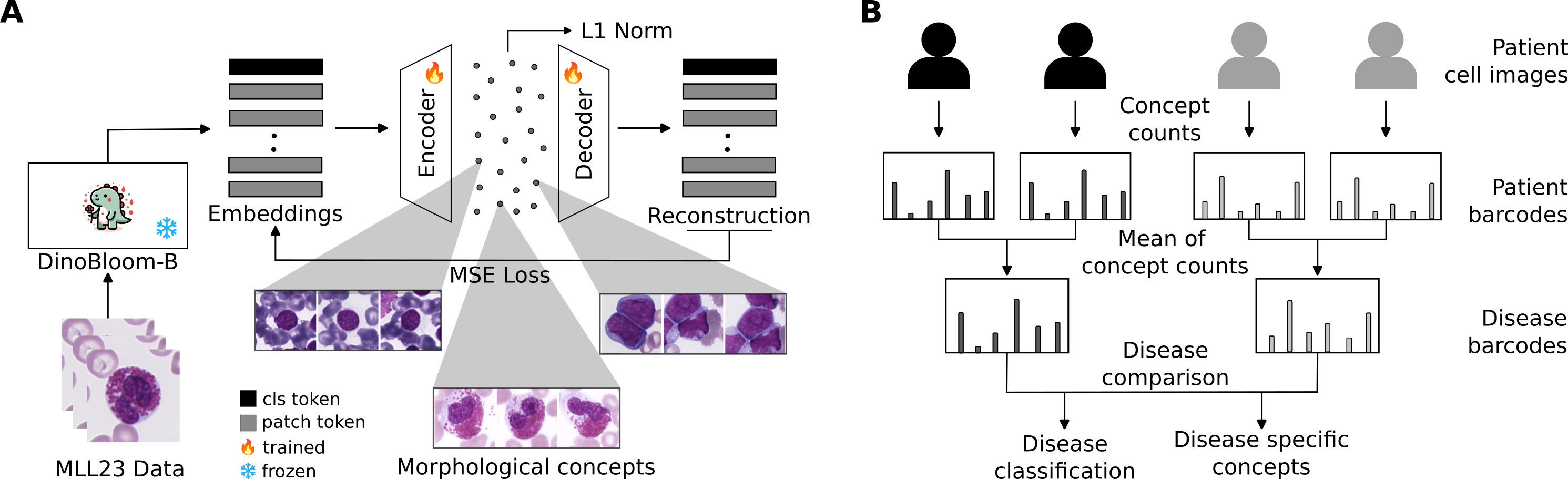}
\captionsetup{labelformat=empty}
\caption{\textbf{Fig. 1. Overview. (A)} CytoSAE is trained on over 40K peripheral blood single-cell images using embeddings from DinoBloom-B, disentangling compact embedding information into meaningful concepts. We employ MSE loss for embedding reconstruction and apply L1 normalization to SAE latents. \textbf{(B)} Patient-level information is aggregated by counting the number of activations for each concept within the patent's single-cell images, forming a ``barcode''. This analysis is extended to the disease level to identify disease-specific morphological concepts.}
\label{fig1}
\end{figure}

CytoSAE is a two-layer network trained with a reconstruction objective. A token from the input model is mapped to a high-dimensional latent space using a linear layer and non-linearity, then mapped back to the input space with a linear layer, 
\begin{align*}
    z &= f(x) =  \textrm{ReLU}(W_{enc}(x-b_{dec})+b_{enc}), \\
    \hat{x} &= g(z) = W_{dec}z + b_{dec},
\end{align*}
where \(x\in \mathbb R^{d_m}, W_{enc}\in \mathbb R^{d_{\text{SAE}}\times d_m}, W_{dec}\in \mathbb R^{d_m\times d_{\text{SAE}}}, b_{enc}\in \mathbb R^{d_{\text{SAE}}},\) and \(b_{dec}\in \mathbb R^{d_m}\).
We optimize the mean squared error (MSE) for token reconstruction while enforcing sparsity in its latent representation through an L1 norm regularizer~\cite{bricken2023monosemanticity},
\begin{equation}
    \min_{W_{enc}, b_{enc}, W_{dec}, b_{dec}}\sum_{i = 1}^K 
    \left(
        \Vert x_i - \hat{x}_i \Vert^2_2 + \lambda \Vert f(x_i)\Vert_1
    \right),
\end{equation}
for tokens $x_1,\dots,x_K$ generated by the backbone model.
We center the input by subtracting the decoder bias $b_{dec}$ initialized with the geometric median of the training tokens. We use ghost gradient resampling \cite{bricken2023monosemanticity} to limit unused latents.

\subsubsection{Training and validation.}
    We obtained tokens from DinoBloom-B \cite{koch2024dinobloom}, a hematology foundation model that partitions images into $14\times14$ sized patches. It processes 257 tokens, including the CLS token, each with a dimension of 768. For the SAE, we expanded this dimension by a factor of 64 and arrived at a 49,152 dimensional latent space. Among various options for backbone model representations to decompose, we used the residual stream output from the second last attention layer. 
    After feeding cell images into DinoBloom-B, we extracted the residual layer output and passed it to the SAE. 
    During model development, we performed an ablation study on the hyperparameters and choose the value highlighted with \(*\): expansion factor [16, 32, 64\(^*\), 128], \(b_{dec}\) initialization method [mean, geometric median\(^*\)], residual output layer [2, 5, 11\(^*\), 12], ghost gradient resampling [True\(^*\), False], L1 coefficient [\(8\times 10^{-4}\), \(8\times 10^{-5}\)\(^*\), \(8\times 10^{-6}\)], and learning rate [\(4\times 10^{-3}\), \(4\times 10^{-4}\)\(^*\), \(4\times 10^{-5}\)] with a constant warmup schedule (500 warmup steps). We used MSE, L1, and L0 loss to measure training performance.
    Each experiment was repeated three times with different random seeds. 

    We trained CytoSAE on the MLL23 dataset \cite{shetab2025large}, which includes 41,906 peripheral blood single-cell images from 18 different cell types (Fig.~\ref{fig1}A).
    The dataset includes blood cells at various maturation stages and abnormalities from both the myeloid and lymphoid lineages. 
    After training, we validated the CytoSAE model on various datasets, encompassing both peripheral blood and bone marrow cytology.

\subsubsection{Analysis and evaluations.}
    We evaluate CytoSAE on MLL23 and four additional datasets.
    {Acevedo} \cite{acevedo2020dataset} contains 17,092 single-cell images labeled into 11 classes.
    {Matek19} \cite{Matek2019} consists of 18,365 expert-labeled single-cell images from peripheral blood smears, classified into 15 classes.
    {BMC} \cite{Matek2021} includes 171,373 expert-annotated cells from bone marrow smears.
    {AML Hehr} \cite{hehr2023explainable} contains patient-wise arranged single-cell images from 189 patients, covering four genetic AML subtypes and a healthy control group.

    To identify morphological concepts, we first collected maximally activated reference images for each SAE latent from all five datasets and collaborated with an expert cytomorphologist with approx. 15 years of experience for validation.
    We randomly sampled 50 latents from high-activation clusters (above a mean threshold of -3) using the KMeans clustering algorithm with \( k=10 \), selecting 5 latents from each cluster. The expert reviewed and annotated the latents whether they represent meaningful concepts.
    Patch-wise analysis enables a subcellular understanding of morphological concepts and allows for validation across different datasets. To derive a global interpretation of an image for a given SAE latent, we aggregated patch-level activations into an image-level activation by counting the number of patches that activate the corresponding latent \cite{lim2024sparse}.

    Specifically, we first binarize the SAE latent activation at the \( j \)-th patch of image \( i \) for the \( s \)-th latent using a threshold \( \tau \) (\(a_{i,j}[s]\)). From this statistic, we derive image level ($a_i$), patient-level ($a_p$) and disease-level ($a_d$) features:
    \begin{align}
        a_{i,j}[s] &= I(h_{i,j}[s] > \tau), \label{Eq1} \\
        a_i[s] = \sum_{j=1}^{n_i} a_{i,j}[s],~
        a_p[s] &= \frac{1}{N_p} \sum_{i \in I_p} a_i[s],~
        a_d[s] = \frac{1}{N_d} \sum_{p \in P_d} a_p[s],
        \label{Eq2}
    \end{align}
    where \( I(\cdot) \) is an indicator function,  \( I_p \) represents the set of images associated with patient \( p \), and \( P_d \) represents the set of patients diagnosed with disease \( d \).
    This hierarchical aggregation allows the identification of shared morphological features at the patient and disease levels while accounting for variations in image and patient count.
    
\subsubsection{Disease Classification.}  
    Using the AML Hehr dataset, we generated patient barcodes \( a_p \) (Eq.~\ref{Eq2}). To assess the predictive power of barcodes and encoded concepts, we applied linear probing to quantify the intermediate representations captured by the SAE. Specifically, we attempted to classify patient disease status using only barcodes. For linear probing, we used logistic regression with L2 regularization. The regularization coefficient was set to \( \frac{c \times n}{100} \), where \( n \) is the number of training samples and \( c \) corresponds to the number of classes \cite{koch2024dinobloom}.

\section{Results}

\begin{figure}[t]
\includegraphics[width=\textwidth]{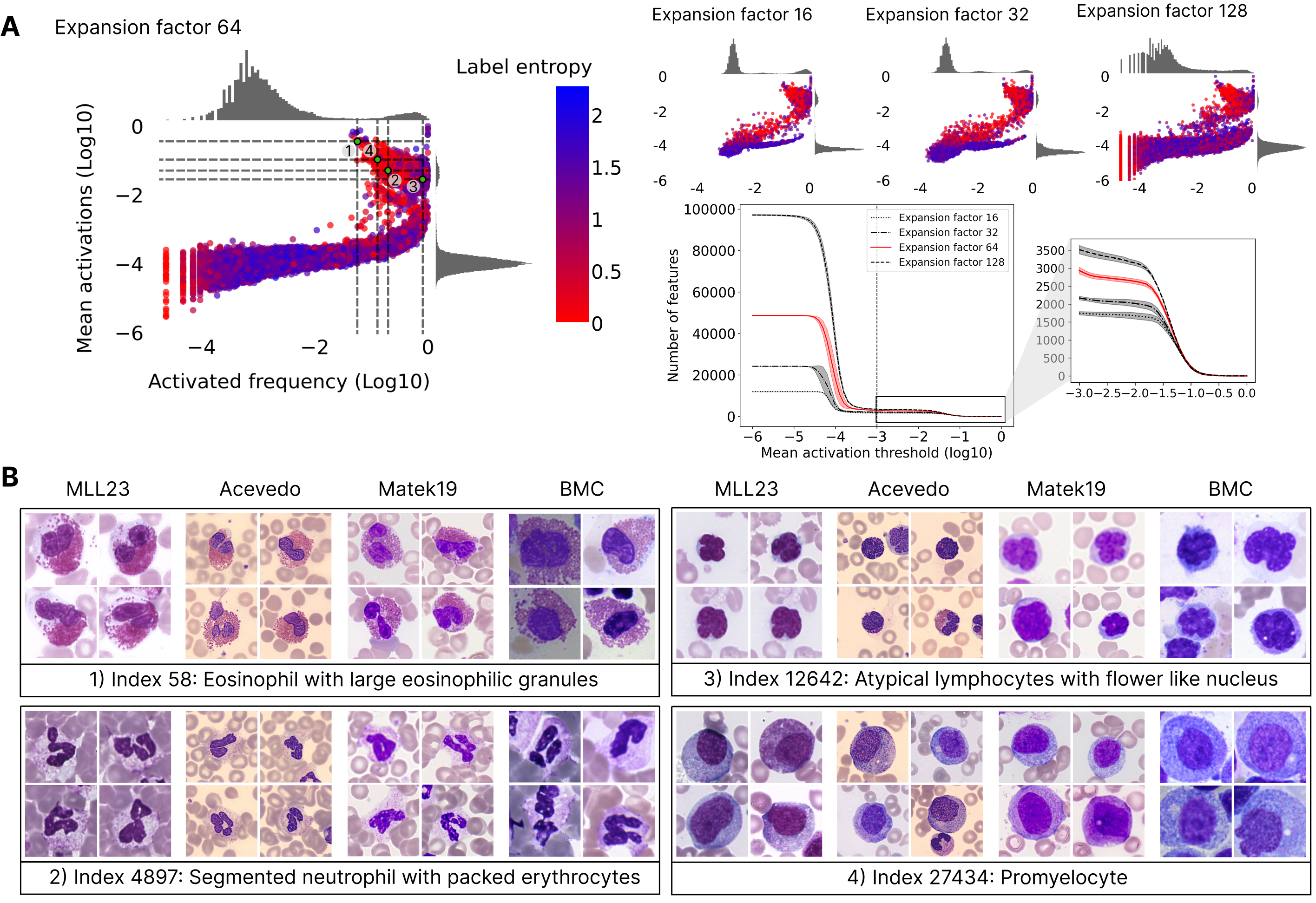}
\captionsetup{labelformat=empty}
\caption{\textbf{Fig. 2. CytoSAE discovers morphologically relevant concepts across datasets. (A)} Scatter plot of SAE latents with varying expansion factors (x-axis: log$_{10}$ of activation frequency, y-axis: log$_{10}$ of mean activation). Latents are color-coded by label entropy. Histograms show latent distributions along corresponding axes. Bottom right: Comparison of the number of latents exceeding a given threshold across models with different SAE dimensions. \textbf{(B)} Reference images from four datasets corresponding to selected latents.}\label{fig2}
\end{figure}

\subsection{CytoSAE discovers morphologically relevant concepts}
    First, we evaluated the effect of different expansion factors on the latent distribution. Specifically, we analyzed the mean latent activations and their activation frequencies on the MLL23 dataset. The latent distribution is depicted in Fig.~\ref{fig2}A, revealing two distinct clusters of activation patterns. We focused on latents where the model was confident about the underlying concept, i.e., those with high mean activation. To quantify this, we counted the number of latents exceeding various mean activation thresholds.
    Higher-dimensional SAEs produced more latents above any given threshold (Fig.~\ref{fig2}B). However, while the difference was substantial at lower thresholds, the gap became less pronounced at higher thresholds (e.g., at threshold -3, the number of activated latents was 1748±56, 2170±61, 2936±114, and 3518±145 for expansion factors 16, 32, 64, and 128, respectively). Notably, the number of latents in the highly activated cluster—those of primary interest—remained relatively stable (Fig.~\ref{fig2}A, bottom right). 
    Next, we randomly selected latents from the high mean activation cluster and visualized the recovered concepts across the four evaluation datasets (Fig.~\ref{fig2}B). The extracted concepts were consistent across datasets and qualitatively, were not affected by domain shift. Expert manual labeling of these concepts (see Sec.~\ref{sec:2}) confirmed that CytoSAE successfully captured various morphologically relevant features.

    \begin{figure}[t]
    \includegraphics[width=\textwidth]{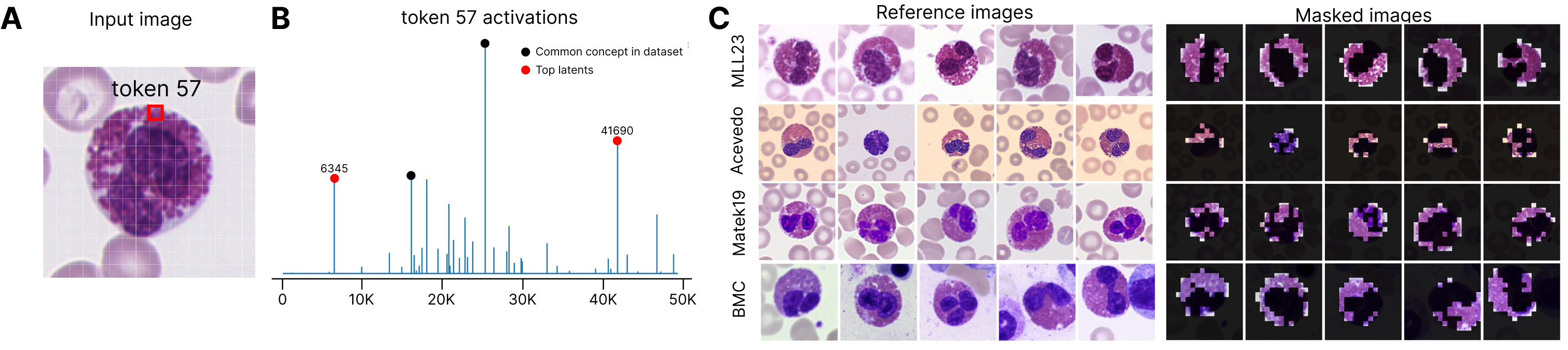}
    \captionsetup{labelformat=empty}
    \caption{\textbf{Fig. 3. Patch-level morphological components are consistent  across diverse datasets. (A)} We highlighted patch 57 of an example image, containing eosinophilic granules, and fed its corresponding token into the SAE. \textbf{(B)} Activated latents of token 57, along with the top two latents after filtering out those shared by all images in the dataset. \textbf{(C)} Reference images with their corresponding patches that activate the selected latents.}
    \label{fig3}
    \end{figure}

\subsection{Patch-wise attribution allows analysis on sub-cellular level}
    Next, we leveraged CytoSAE for patch-level attributions of the discovered concepts. In Fig.~\ref{fig3}, we show the activated latents for an example patch containing eosinophilic granules and selected the top two latents after filtering out those that were activated across the entire dataset.
    Next, we retrieved reference images that also activated these selected latents. Using patch-level latent activations from the retrieved images, we generated segmentation maps that highlighted the same concept present in the given input patch. Specifically, for a given image $x_i$ and an SAE latent index $s$, we visualized the concept by multiplying each patch $x_{i,j}$ with its corresponding latent activation value $h_{i,j}[s]$.  
    Notably, all retrieved images--including those from different datasets and bone marrow cytology--consistently highlighted eosinophilic granules, demonstrating that the latents successfully capture meaningful morphological concepts.  
    
\begin{figure}[t]
\includegraphics[width=\textwidth]{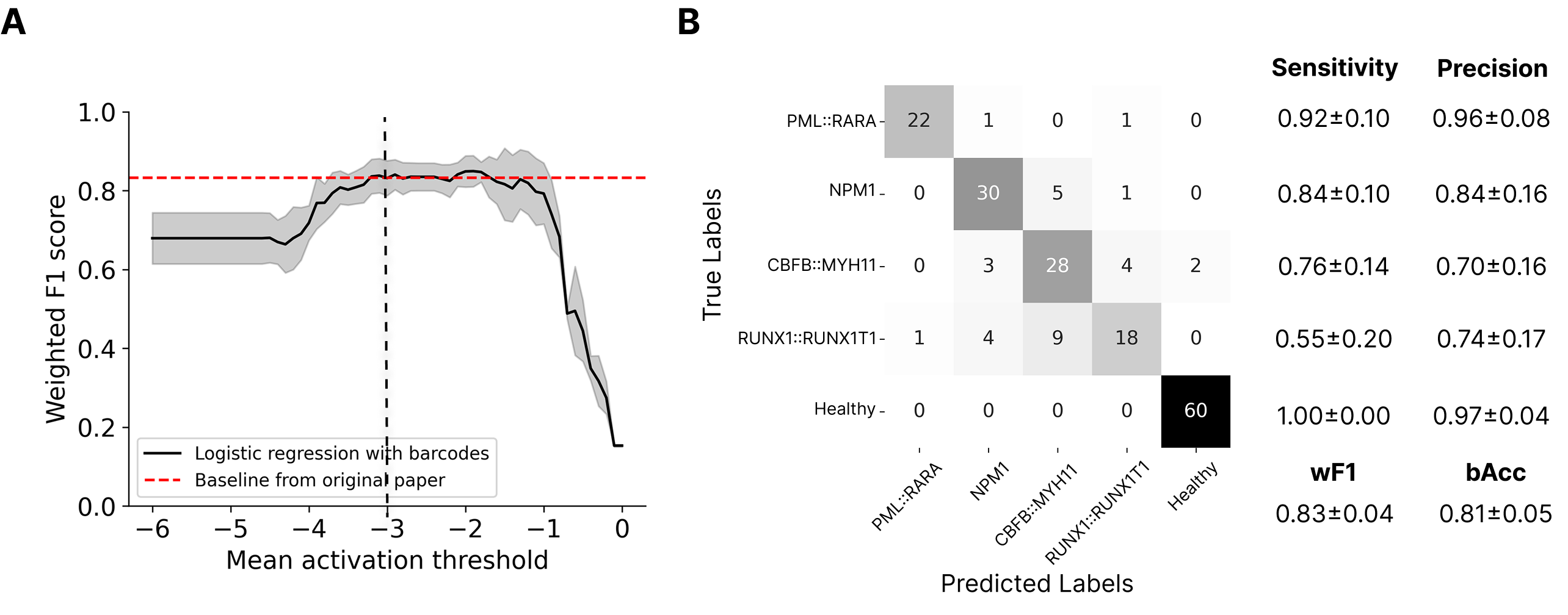}
\captionsetup{labelformat=empty}
\caption{\textbf{Fig. 4. Patient-wise concepts can predict disease. (A)} We train a logistic regression classifier on patient barcodes, varying the number of latents; latents above the threshold were used in classification, with others set to zero.  
\textbf{(B)} Multi-class classification performance of logistic regression using patient barcodes is comparable to a fully deep learning approach \cite{hehr2023explainable}.}
\label{fig4}
\end{figure}

\begin{figure}[t]
\includegraphics[width=\textwidth]{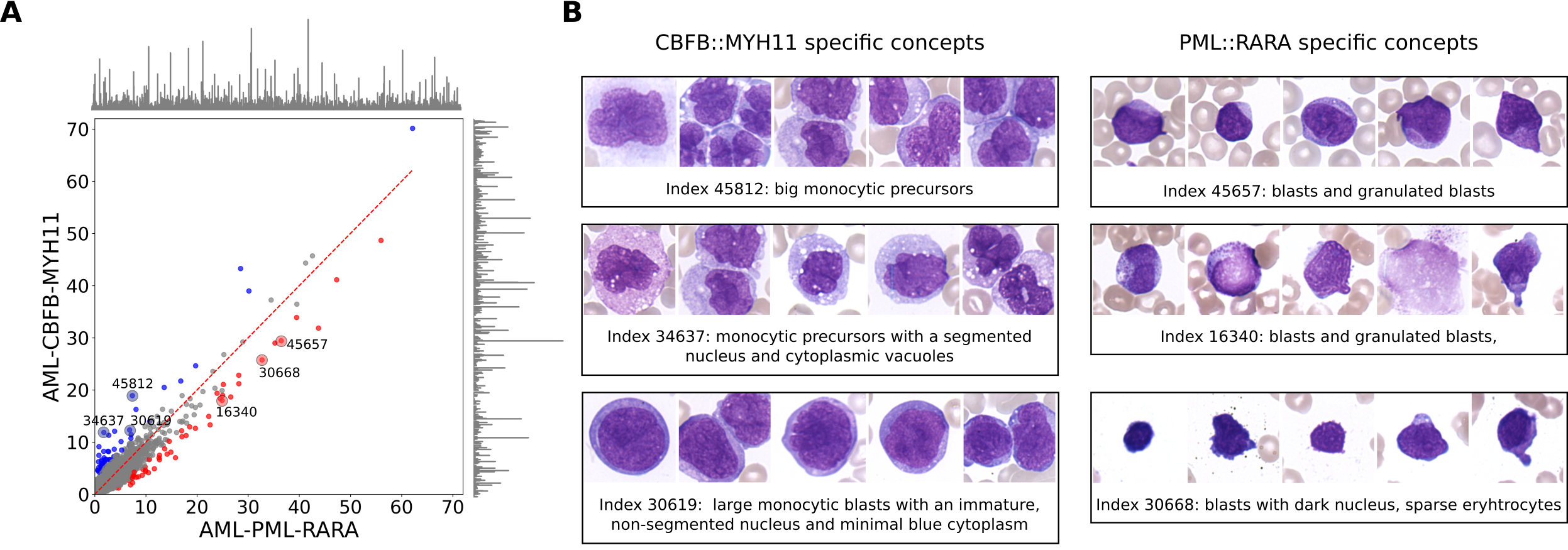}
\captionsetup{labelformat=empty}
\caption{\textbf{Fig. 5. Disease-wise comparison reveals disease-specific morphological features. (A)} Mean count of latents across patients. Blue: top 50 latents that are more frequently activated in the \textit{CBFB::MYH11} subtype; red: top 50 latents for \textit{PML::RARA} subtype. Disease barcodes are positioned along the corresponding axes.\textbf{(B)} Representative images of disease-specific concepts.}
\label{fig5}
\end{figure}

\subsection{Patient-wise analysis discovers disease specific morphologies}
    For patient-wise analysis, we used AML peripheral blood single-cell image data \cite{hehr2023explainable}. This dataset includes four subtypes of AML: (i) APL with \textit{PML::RARA} fusion, (ii) AML with \textit{NPM1} mutation, (iii) AML with \textit{CBFB::MYH11} fusion (without NPM1 mutation), and (iv) AML with \textit{RUNX1::RUNX1T1} fusion, along with healthy stem cell donors.
    For each patient, we counted how many times a latent (morphological concept) was activated across single-cell images and generated a patient-specific barcode. By averaging activation counts across patients with the same disease, we created disease barcodes (Fig.~\ref{fig1}B, Eq.~\ref{Eq2}). This approach enabled both patient-wise comparisons and disease classification while identifying morphological concepts uniquely expressed in individual patients.
    To evaluate the discriminative power of these morphological concepts, we performed disease classification using patient barcodes while varying the number of latents included in the classification (Fig.~\ref{fig4}A). Although adjusting the threshold from -6 to 0 (log$_{10}$) reduced the number of latents used in classification (Fig.~\ref{fig2}A), classification performance remained stable up to a certain threshold and using only meaningful concepts resulted in higher classification performance (Fig.~\ref{fig4}A). Our findings confirm that classification performance is only affected when meaningful concepts are removed. Multi-class classification at a threshold of -3 (log$_{10}$) achieved a weighted F1-score of 0.832 ± 0.044 which is similar to the performance of fully deep learning baseline \cite{hehr2023explainable}.
    By comparing disease barcodes, we identified the top-100 latents that were differentially expressed between the \textit{CBFB::MYH11} and \textit{PML::RARA} subtypes (50 for each disease). Experts annotated these concepts and assessed whether they represented disease-specific morphological features recognizable by a cytomorphologist. Among the top features, 5 out of 10 and 32 out of 50 were identified as \textit{CBFB::MYH11}-specific, while 4 out of 10 and 10 out of 50 were classified as \textit{PML::RARA}-specific.
    \textit{CBFB::MYH11}-specific concepts predominantly highlighted large monocytic precursors with segmented nuclei and cytoplasmic vacuoles (Fig.~\ref{fig5}B). In contrast, \textit{PML::RARA}-specific concepts were characterized by granulated blasts and morphological features suggestive of anemia (Fig.~\ref{fig5}B).

\subsection{Ablation and variation studies}

    With our final set of parameters 2936±114 concepts are discovered at a threshold of -3. Varying the expansion factor changes the number of concepts as this treshold in a range of 1748±56 (expansion 16) to 3518±145 (expansion 128), remaining within the same order of magnitude. Increasing the sparsity regularizer $\lambda$ by 10x results in too few concepts (128±83) at low L0, while lowering by 10x results in a two orders of magnitude increase in the L0, motivating our current choice. Increasing the learning rate too much resulted in unstable training, while lowering to 4e-5 did not meaningfully impact the results, yielding 2755±108 concepts.
    We then varied the number of layers: At layer 2 and 7, we discovered 85±4 and 1253±34 concepts while in the last layer 12, 15161±3477 concepts crossed the threshold. This indicates that the layers between 7--11 are the most interesting choices for finding a good balance in the number of concepts.
    Ghost gradient regularization did not substantially influced number of concepts (2907±133) but decreased the fraction of dead features from 0.92±0.03 to 0.57±0.03.

\section{Conclusion}
In this study, we introduced CytoSAE for morphological concept discovery in hematological cytology. Our findings demonstrate that CytoSAE successfully learns interpretable morphological features across different datasets, including peripheral blood smears and bone marrow cytology, while maintaining generalizability to out-of-domain datasets. This capability is crucial in medical AI, where real-world applications often require robustness to dataset shifts caused by variations in staining, background artifacts, and imaging protocols.
We show patch-level, image-level, patient-level, and disease-level analysis and demonstrated interpretable disease-level subtype classification.

While CytoSAE provides a powerful approach for morphological feature discovery, some limitations remain. Manual expert validation was conducted at the dataset level, but we did not evaluate reference images within each individual patient to avoid human selection bias in concept discovery. Future work could explore patient-specific concept retrieval, though care must be taken to minimize subjectivity in the selection process.
Additionally, future studies could incorporate semi-supervised learning techniques to refine the discovered features further.

\begin{credits}

    \subsubsection{Author contributions.} 
    Conceptualization: StS, CM;
    Methodology: MFD, HL, StS;
    Software: MFD, HL;
    Investigation: MFD;
    Data Curation: MB, KSG, MFD;
    Writing--Original Draft: HL, MFD;
    Writing--Editing: StS, HL, CM.

    \subsubsection{\ackname}
        C.M. acknowledges funding from the European Research Council (ERC) under the European Union's Horizon 2020 research and innovation program (Grant Agreement No. 866411 \& 101113551) and support from the Hightech Agenda Bayern.
    
    \subsubsection{\discintname}
        The authors have no competing interests to declare that are relevant to the content of this article.
        
\end{credits}

\bibliographystyle{splncs04}
\bibliography{references}

\end{document}